# Why Cannot Large Language Models Ever Make True Correct Reasoning?

**Jingde Cheng**

(Professor Emeritus)

Department of Information and Computer Sciences, Saitama University

Saitama, Japan

jingde.cheng@gmail.com, cheng@aise.ics.saitama-u.ac.jp

**Abstract** — Recently, with the application progress of AIGC tools based on large language models (LLMs), led by ChatGPT, many AI experts and more non-professionals are trumpeting the "reasoning ability" of the LLMs. The present author considers that the so-called "reasoning ability" of LLMs are just illusions of those people who with vague concepts. In fact, the LLMs can never have the true reasoning ability. This paper intents to explain that, because the essential limitations of their working principle, the LLMs can never have the ability of true correct reasoning.

## I. Introduction

Recently, with the application progress of AIGC tools based on large language models (LLMs), led by ChatGPT, many AI experts and more non-professionals are trumpeting the "reasoning ability" of the LLMs. The present author considers that the so-called "reasoning ability" of LLMs are just illusions of those people who with vague concepts. In fact, due to the limitations of working principle, the LLMs can never have the ability of true correct reasoning.

After defining what is correct reasoning strictly and explicitly, and discussing the logical basis to underlie correct reasoning, this paper intents to explain that, because the essential limitations of their working principle, the LLMs can never have the ability of true correct reasoning. Seeking the true correct reasoning ability of LLMs is no different from seeking perpetual motion machines; seeking the reasoning ability of LLMs but without consideration on its correctness evaluation criterion is a completely wrong and hopeless research direction.

The rest of the paper is organized as follows: Section 2 defines basic notions concerning correct reasoning, Section 3 discusses the logical basis for correct reasoning, Section 4 shows why LLMs can never have the ability of true correct reasoning, and concluding remarks are given in Section 5.

## II. What is Correct Reasoning?

***Reasoning*** is an ordered process of drawing new conclusion from given premises, which are already known facts or previously assumed hypotheses to provide conclusive relevant evidence for the conclusion. In general, a reasoning is composed of multiple arguments in a certain order.

About our definition of reasoning, there are two important points that must be mentioned as follows:

First, we directly take "relevance" between the premises and the conclusion into the definition of reasoning; this is intrinsically different from those definitions of reasoning in all traditional textbooks of logic, dictionaries, and handbooks [e.g., 1-6]. We will discuss this matter in the next section.

Second, our definition of reasoning is constructive and functional but not mental/psychological: it just shows how a reasoning is composed and what is its function/goal, but does not concern how a reasoning is made and who is the subject of behavior.

Note that in the definition of reasoning, the premises of a reasoning are supposed to provide conclusive relevant evidence for the conclusion of that reasoning. However, though the premises of a reasoning are intended to provide conclusive relevant evidence for the conclusion of that reasoning, they not necessarily actually do so.

From the perspective of composition, any reasoning is composed of three elements: some given premises, a process consisting of ordered drawing steps (i.e., arguments), and a conclusion. Therefore, anything with these three elements can be regarded as a kind of reasoning, but whether the reasoning is really correct or reasonable is another matter completely. Only those reasoning whose premises really provide conclusive relevant evidence for the new conclusion are correct and reasonable.

An *argument* is a set of propositions or statements (declarative sentences) of which one proposition/statement is intended as the "conclusion", and the other propositions/statements, called "premises," are intended to provide conclusive relevant evidence for the conclusion.

Again, note that we directly take "relevance" between the premises and the conclusion into the definition of argument; this is intrinsically different from those definitions of argument in all traditional textbooks of logic, dictionaries, and handbooks [e.g., 1-6].

Therefore, any set of propositions or statements can be regarded as an argument, but whether the argument is really correct is another matter completely. Only those arguments which premises really provide conclusive relevant evidence for the conclusion are correct.

There are three types of arguments: deductive arguments, inductive arguments, and abductive arguments [2-6].

A *deductive argument* is an argument in which the premises with some general propositions/statements are intended to provide absolute support (evidence) for the specific conclusion, i.e., its conclusion follows necessarily from its premises.

A reasoning is called a "*deductive reasoning*" if its every argument is a deductive argument.

An example of deductive argument (Modus Ponens): If A then B, A, therefore B. Here, "If A then B" is a general proposition/statement, "B" is a specific proposition/statement, and "B" follows necessarily from premises "If A then B" and "A".

An *inductive argument* is an argument in which the special/particular premises are intended to provide some degree of support (evidence) for the general conclusion, i.e., its conclusion is not necessarily follows from its premises.

A reasoning is called an "*inductive reasoning*" if its every argument is an inductive argument.

An example of inductive argument: If $A_1$ is a B, $A_2$ is a B, …, $A_n$ is a B, then maybe all $A_1$, $A_2$, …, $A_n$, $A_{n+1}$, $A_{n+2}$, …, are also B.

An *abductive argument* (also often called *Inference to the Best Explanation, IBE*) [7-9] is an argument in which the premises are intended to provide a hypothesis as the conclusion.

A reasoning is called an "*abductive reasoning*" if its every argument is an abductive argument.

An example of abductive argument: If A then C, C, therefore maybe A**.** ("The surprising fact, C, is observed.  But if A were true, C would be a matter of course. Hence, there is reason to suspect that A is true".) [C. S. Peirce]

Note that in the all three types of arguments, the notion of **conditional**, "if … then …", plays a core role.  We will discuss it in the next section.

In general, a complex reasoning process may include all three types of arguments.

An argument is a conclusion standing in relation to its supporting evidence.  In an argument, a claim is being made that there is some sort (absolute or some degree) of evidential relation between its premises and its conclusion: the conclusion is supposed to follow from the premises, or equivalently, the premises are supposed to entail the conclusion.

Therefore, the correctness of an argument is a matter of the connection between its premises and its conclusion, and concerns the strength of the relevant relation between them.  Note that the correctness of an argument depends neither on whether the premises are really true or not, nor on whether the conclusion is really true or not.

Because we define that only those arguments/reasoning whose premises really provide conclusive relevant evidence for the conclusions are correct or reasonable, and in different type of arguments/reasoning, the premises provide different kinds of evidence (absolute or some degree) for the conclusion, so we must distinguish the correctness of arguments/reasoning into different types.

For a deductive argument/reasoning, it is **valid** if its premises really provide conclusive relevant evidence for its conclusion, otherwise, it is **invalid**.

Here we directly take "relevance" between the premises and the conclusion into account of validity, this is intrinsically different from those definitions of validity in all traditional textbooks of logic.  We will discuss this in the next section.

Therefore, a valid deductive argument is one whose conclusion is claimed to follow from its premises with absolute necessity, this necessity not being a matter of degree and not depending in any way on whatever else may be the case.

For an inductive argument/reasoning, it is **strong** if in which the conclusion does in fact follow probably from the premises; it is **weak** if in which the conclusion does not follow probably from the premises, even though it is claimed to.

An inductive argument claims that its premises give only some degree of probability, but not certainty, to its conclusion.

Therefore, an inductive argument is one whose conclusion is claimed to follow from its premises only with probability, this probability being a matter of degree and dependent on what else may be the case.

For an abductive argument/reasoning, its correctness means explanatory goodness -- how well the hypothesis explains the data compared to alternatives.  Harman defines abductive correctness in terms of choosing the hypothesis that best explains the evidence according to criteria like simplicity and coherence.

From the perspective of **correctness**, the premises of any correct reasoning must provide conclusive relevant evidence for the conclusion.

From the perspective of **effectiveness**, the conclusion of any effective reasoning must be new, i.e., it is certainly need to be drawn from the premises.

Only **correct** and **effective** reasoning is **reasonable**, and nothing else is reasonable.

Therefore, any correct reasoning is intrinsically ampliative, i.e., it has the function of enlarging or extending some things, or adding some new things to what is already known or assumed.

Note that how to define the notion of "new" formally and satisfactorily is still a difficult open problem until now. Its discussion is beyond this paper.

Let us see two examples of reasoning:

(1) If a number is a rational number, then it must be expressed as the ratio of a pair of integers.

(2) $\pi$ cannot be expressed as the ratio of a pair of integers.

Therefore,

    (3) $\pi$ is not a rational number.

    (4) $\pi$ is a number.

Therefore,

    (5) At least one irrational number exists.

Here, we use one argument to deduce (3) from premises (1) and (2) as the conclusion, and one argument to deduce (5) from premises (3) and (4) as the conclusion. Such two-step argument constitutes a complete reasoning, and a new conclusion (5) is deduced from premises (1), (2) and (4). This is a correct reasoning.

However, if we replace (4) by

(4') e is a number.

then it is not a correct reasoning that from premises (1), (2) and (4') to deduce (5) as a new conclusion, because (3) and (4') cannot provide conclusive relevant evidence for (5), even if (5) does not changed and still a fact.

Therefore, there are correct reasoning and incorrect reasoning.

Thus, there are some fundamental questions: How to distinguish correct reasoning from incorrect reasoning? What is the evaluation criterion by which one can decide whether the conclusion of an argument or a reasoning really does follow from its premises or not? Is there the only one criterion, or are there many criteria? If there are many criteria, what are the intrinsic differences between them?

It is logic, as a fundamental discipline, that deals with these questions from various philosophical viewpoints.

## III. LOGICAL BASIS FOR CORRECT REASONING

"Logic is the study of the methods and principles used to distinguish good (correct) from bad (incorrect) reasoning." "The study of logic aims to discover and make available those criteria that can be used to test arguments for correctness." "The distinction between correct and incorrect reasoning is the central problem with which logic deals. The logician's methods and techniques have been developed primarily for the purpose of making this distinction clear." [2]

*Logic* deals with what entails what or what follows from what, and aims at determining which are the correct conclusions of a given set of premises, i.e., to determine which arguments are correct and/or valid. Therefore, the most essential and central concept in logic is the ***logical consequence relation*** that relates a given set of premises to those conclusions, which validly follow from the premises. To define a logical consequence relation is nothing else but to provide a logical validity criterion by which one can decide whether the conclusion of an argument or a reasoning really does follow from its premises or not. Moreover, to answer the question what is the correct conclusion of given premises, we have to answer the question: correct for what? Based on different philosophical motivations, one can define various logical consequence relations and therefore construct various logic systems.

A ***logically valid reasoning*** is a reasoning such that its arguments are justified based on some logical validity criterion provided by a logic system in order to obtain correct conclusions (Note that here the term 'correct' does not necessarily mean 'true'). Therefore, the logical validity of reasoning is dependent on the logic system adapted as the validity criterion, a reasoning is valid according to a logic but may be invalid according to another logic. To say a reasoning is valid but does not explicitly specify the logical validity criterion is nonsense.

In logic, a sentence in the form of "if ... then ..." is usually called a ***conditional proposition*** or simply ***conditional*** which states that there exists a relation of sufficient condition between the "if" part and the "then" part of the sentence. In general, a conditional must concern two parts which are connected by the connective "if ... then ..." and called the ***antecedent*** and the ***consequent*** of that conditional, respectively. The truth of a conditional depends not only on the truth of its antecedent and consequent but also, and more essentially, on a necessarily relevant and conditional relation between them. The notion of conditional plays the most essential role in reasoning because any reasoning form must invoke it, and therefore, it is historically always the most important subject studied in logic and is regarded as "the heart of logic" [10]. Based on different philosophical motivations, logicians have defined different logical connectives to represent the notion of conditional and then constructed various different logic systems. For a logic system, a conditional is called an ***entailment*** if it is a logical theorem of that logic system. "The problem in modern logic can best be put as follows: can we give an explanation of those conditionals that represent an entailment relation?" [11]

Now, our problem is which logic system can underlie correct reasoning satisfactorily?

The present author considers that the fundamental logic system underlying correct reasoning must satisfy the following three essential requirements [12,13]:

First, as a general logical criterion for the validity of reasoning, the fundamental logic must be able to underlie correct reasoning as well as truth-preserving reasoning in the sense of conditional. Recall our definition of reasoning, for any reasoning to be correct, it is a necessary condition that its premises must provide conclusive relevant evidence for the conclusion. On the other hand, for any reasoning based on the logic to be valid, if its premises are true in the sense of conditional, then its conclusion also must be true in the sense of conditional.

Logic is the study of the methods and principles used to distinguish correct reasoning from incorrect reasoning, and the notion of a conditional is the heart of logic. Because there is no reasoning that does not invoke the notion of conditional, this requirement is primarily important.

Second, the fundamental logic must be able to underlie ampliative reasoning, i.e., for any reasoning based on the logic to be valid, the truth of conclusion of the reasoning should be recognized after the completion of the reasoning process but not be invoked in deciding the truth of premises of the reasoning.

From the perspective of regarding reasoning as the process of drawing new conclusions from given premises, any meaningful and effective reasoning must be ampliative but not circular and/or tautological.

Third, the fundamental logic must be able to underlie paracomplete reasoning and paraconsistent reasoning, i.e., the conclusion of a reasoning may not be the negation of a sentence even if the sentence is not a conclusion of the premises of that reasoning, and also may not be an arbitrary sentence even if the premises of that reasoning is inconsistent. In particular, the so-called principle of Explosion that everything follows from a contradiction should not be accepted by the logic as a valid principle.

In general, our knowledge about a domain as well as a scientific discipline may be incomplete and/or inconsistent in many ways, i.e., it gives us no evidence for deciding the truth of either a proposition or its negation, and/or it directly or indirectly includes some contradictions. Therefore, reasoning with incomplete and/or inconsistent knowledge is the rule rather than the exception in our everyday lives and almost all scientific disciplines.

ONLY IF the fundamental logic system satisfies the above three essential requirements, for any valid reasoning with true premises based on the fundamental logic system, we can directly accept its conclusion as correct one without evaluation.

Classical mathematical logic (CML) was established in order to provide formal languages for describing the structures with which mathematicians work, and the methods of proof available to them; its principal aim is a precise and adequate understanding of the notion of mathematical proof. CML was established based on a number of fundamental assumptions. Among them, the most characteristic one is the classical account of validity (CAV for short) that is the logical validity criterion of CML by which one can decide whether the conclusion of an argument or a reasoning really does follow from its premises or not in the framework of CML. However, since the relevance between the premises and conclusion of an argument is not accounted for by the CAV (i.e., an argument/reasoning is classically valid if and only if it is impossible for all its premises to be true while its conclusion is false), a reasoning based on CML is not necessarily relevant. On the other hand, in CML the notion of conditional, which is intrinsically intensional but not truth-functional, is represented by the notion of material implication, which is intrinsically an extensional truth-function. This leads to the problem of implicational paradoxes [10-13].

CML cannot satisfy any of the above three essential requirements for the fundamental logic system because of the following facts: a reasoning based on CML is not necessarily relevant; the classical truth-preserving property of a reasoning based on CML is meaningless in the sense of conditional; a reasoning based on CML must be circular and/or tautological but not ampliative; reasoning under inconsistency is impossible within the framework of CML [12,13].

The above facts are also true to ALL classical conservative extensions or non-classical alternatives of CML where the CAV is adopted as the logical validity criterion and the notion of conditional is directly or indirectly represented by the material implication [12,13].

Traditional relevant (or relevance) logics (RLs) were constructed during the 1950s in order to find a mathematically satisfactory way of grasping the elusive notion of relevance of antecedent to consequent in conditionals, and to obtain a notion of implication which is free from the so-called paradoxes of material and strict implication [10,11]. Some major traditional RL systems are 'system E of entailment', 'system R of relevant implication', and 'system T of ticket entailment'. A major characteristic of the RLs is that they have a primitive intensional connective to represent the notion of conditional (entailment) and their logical theorems include no implicational paradoxes. The underlying principle of the RLs is the relevance principle, i.e., for any entailment provable in E, R, or T, its antecedent and consequent must share a propositional variable. Variable-sharing is a formal notion designed to reflect the idea that there be a meaning-connection between the antecedent and consequent of an entailment. It is this relevance principle that excludes those implicational paradoxes from logical axioms or theorems of the RLs. Also, since the notion of entailment is represented in the RLs by a primitive intensional connective but not an extensional truth-function, a reasoning based on the RLs is ampliative but not circular and/or tautological. Moreover, because the RLs reject the principle of Explosion, they can certainly underlie paraconsistent reasoning [10-13].

However, because the relevance principle cannot exclude those conditionals whose antecedent including unnecessary and needless conjuncts or whose consequent including unnecessary and needless disjuncts, logical theorems of the RLs still include some conjunction-implicational paradoxes and disjunction-implicational paradoxes [12,13].

In order to establish a satisfactory logic calculus of conditional to underlie relevant reasoning, the present author has proposed some strong relevant (or relevance) logics (SRLs), named Rc, Ec, and Tc [12,13]. The SRLs require that the premises of an argument represented by a conditional include no unnecessary and needless conjuncts and the conclusion of that argument includes no unnecessary and needless disjuncts. As a modification of RLs R, E, and T, SRLs Rc, Ec, and Tc rejects all conjunction-implicational paradoxes and disjunction-implicational paradoxes in R, E, and T, respectively. What underlies the SRLs is the strong relevance principle: for any theorem of Rc, Ec, or Tc, every propositional variable in the theorem occurs at least once as an antecedent part and at least once as a consequent part. Since the SRLs are free of not only implicational paradoxes but also conjunction-implicational and disjunction-implicational paradoxes, in the framework of the SRLs, if a reasoning is valid, then both the relevance between its premises and its conclusion and the truth of its conclusion in the sense of conditional can be guaranteed in a certain sense of strong relevance [12,13].

SRLs can satisfy the above three essential requirements for the fundamental logic system to underlie correct reasoning. At present, SRLs is the only family of logics that can satisfy the above three essential requirements.

Therefore, for any theoretical or practical system, if the system provides reasoning facilities for its users, i.e., to reason something for its users or allow users reason something, then it must have a mechanism based on SRLs to make true correct reasoning but rule out incorrect reasoning. In particular, in a crucial application area that cannot allow any incorrect reasoning, this mechanism is most important. Without such a mechanism, any system cannot provide guarantee for its users to get correct results.

Here by "make true correct reasoning" we mean that any reasoning process made by the (machine or human) reasoner(s) must be correct, i.e., the premises of the reasoning must provide conclusive relevant evidence for the conclusion, and therefore, produce 100% logically correct results without incorrect garbage.

### IV. IN-PRINCIPLE LIMITATIONS OF LLMS

First of all, let us see the state of the art of "reasoning ability" of LLMs.

Basically, "LLMs are generative mathematical models of the statistical distribution of tokens in the vast public corpus of human-generated text, where the tokens in question include words, parts of words, or individual characters — including punctuation marks." [14]

Many AI experts and more non-professionals are trumpeting the "reasoning ability" of LLMs. There are many papers, formally published or unpublished but submitted at arXiv, discussed the "reasoning ability" of LLMs.

For example, a resent comprehensive overview of LLMs, which is just accepted for publication by ACM transactions on Intelligent Systems and Technology, claimed: "In addition to better generalization and domain adaptation, LLMs appear to have emergent abilities, such as reasoning, planning, decision-making, in-context learning, answering in zero-shot settings, etc. These abilities are known to be acquired by them due to their gigantic scale even when the pre-trained LLMs are not trained specifically to possess these attributes." [15]

The overview paper also claimed: "Reasoning in LLMs: LLMs are zero-shot reasoners and can be provoked to generate answers to logical problems, task planning, critical thinking, etc. with reasoning. Generating reasons is possible only by using different prompting styles, whereas to improve LLMs further on reasoning tasks many methods train them on reasoning datasets." [15]

However, a basic fact is that the overview paper mentions no word about the correctness (or reasonability, validity) of "reasoning" performed by LLMs, even if the paper enumerated so many "reasoning" facilities in various applications of LLMs. Note that this fact does not simply mean that the paper overlook the correctness evaluation criterion of reasoning, but means that there is no paper in literature discussing the matter, because the paper is a resent overview.

Why do we humans reason? It is because we need and want to get new and correct conclusions by reasoning! Therefore, the correctness evaluation criterion of reasoning is intrinsically important. To discover and make available those correctness evaluation criteria is the main task of logicians. This is why John Duns Scotus said "Logic is the science of sciences, and the art of arts." and other famous logicians, such as Alfred Tarski, Kurt Friedrich Gödel, Ludwig Josef Johann Wittgenstein, said "Logic is the basis for all other sciences."

As we have pointed out in the previous sections, anything with three elements, some given premises, a drawing process, and a conclusion, can be regarded as a kind of reasoning, but whether the reasoning is really correct or reasonable is another matter completely. Only those reasoning whose premises really provide conclusive relevant evidence for the new conclusion are correct and reasonable. Any "reasoning" that is not correct and reasonable is completely useless!

Therefore, although there are so many papers (formally published and unpublished but submitted at arXiv) investigated the "reasoning ability" of LLMs, these investigations all are just illusions of those people who with vague concepts.

Next, let us see why these illusions about the "reasoning ability" of LLMs occurred. The present author considers that LLMs give people the illusion of reasoning ability because of a mix of how humans understand reasoning well and how LLMs are trained and pretended to simulate reasoning well.

First, probably almost people, including a lot of AI experts, saying "reasoning" in fact do not understand what is reasoning, in particular, true correct reasoning. Therefore, when they talk about "reasoning", they do not realize that only true correct reasoning makes sense, just talking about "reasoning" but ignoring its correctness is nonsense.

Second, LLMs are trained on vast amounts of text where humans are reasoning and of course include many good reasoning examples. Therefore, LLMs sometimes (but NOT 100%) may output (copy!) some good reasoning example, and this is mistaken by people for having the reasoning ability by LLMs themselves.

Third, LLMs are trained so powerful such that they can communicate with their human users fluently, and therefore, lead to the ELIZA effect [16]: a tendency to project human traits onto systems having a textual interface. As a result, let people believe that it is the LLM that really makes reasoning when it communicates with humans.

Fourth, LLMs are trained so powerful such that they can simulate (NOT make!) reasoning well enough to solve some problems that are difficult to some human users. As a result, let people believe that the "reasoning ability" of LLMs is higher than their human users, and it is a true correct reasoning ability.

Finally, let us see why LLMs can never have the ability of true correct reasoning.

First of all, recall our definition for "make true correct reasoning": "any reasoning process made by the (machine or human) reasoner(s) must be correct, i.e., the premises of the reasoning must provide conclusive relevant evidence for the conclusion, and therefore, produce 100% logically correct results, without incorrect garbage."

Therefore, this 100% correctness requirement is an insurmountable obstacle for LLMs working based on the fundamental theory combining ideas from probability theory, statistics, and deep learning, because in principle no LLM can produce correct results with 100% certainty.

Since its birth, the computability theory (the theoretical basis of modern electronic digital computers) established in 1930s has always been aimed at "accurate computing", i.e., computing with 100% correctness. Almost applications of computation are also require results as accurate as possible. However, with the application progress of AIGC tools based on LLMs, today, some computing application modes in human society have changed from "accurate computing" to "casual computing" [17].

Thus, is there some possibility that current LLMs will be improved such that they can produce results with a certain degree of correctness? In particular, there are so many AI experts believe and trumpet so-called "scaling laws" for LLMs.

The answer of the present author to the above question is "NO". The key point is that the working principle and architecture of LLMs cannot be embedded a correctness evaluation criterion and a dynamic evaluation mechanism. As long as the LLMs are working based on probability theory, statistics, and deep learning, a formal logic system cannot be embedded as the built-in logical validity evaluation criterion; as long as the LLMs take text tokens as intermediate and final manipulation objects to generate one token at a time in a incremental process style, a dynamic evaluation mechanism from a global viewpoint cannot be embedded. Without a fundamental logic system, any correctness evaluation criterion and dynamic evaluation mechanism for true correct reasoning is impossible.

Another point is: in LLMs, there is no true "truth" and/or "correctness" certainly related to the real world, even if in those good examples copied from human text data; all "truth" and/or "correctness" in LLMs are just "statistical plausibility in text" but completely not correspondence to reality. Therefore, such "statistical plausibility in text" may be OK for some applications, but is meaningless for those applications that require a certain degree of correctness.

Consequently, LLMs can only simulate the form of reasoning (i.e., the three elements) but in principle lack and cannot build-in any correctness evaluation criterion and dynamic evaluation mechanism for true correct reasoning. Therefore, LLMs can never have the ability of true correct reasoning.

## V. Concluding remarks

We have defined what is true correct reasoning strictly and explicitly, presented three essential requirements for the fundamental logic system underlying correct reasoning, showed that strong relevant logic is the only family of logics satisfying the three essential requirements; we also showed that the so-called "reasoning ability" of LLMs is just illusions of those people who with vague concepts, why these illusions about the "reasoning ability" of LLMs occurred, and why LLMs can never have the ability of true correct reasoning.

We can conclude: because the essential limitations of their working principle, the LLMs can never have the ability of true correct reasoning; seeking the reasoning ability of LLMs but without consideration on correctness evaluation criterion is a completely wrong and hopeless research direction.